\documentclass[10pt,twocolumn,letterpaper]{article}

\usepackage[pagenumbers]{cvpr} 

\usepackage{graphicx}
\usepackage{amsmath}
\usepackage{amssymb}
\usepackage{booktabs}
\usepackage{xspace}
\usepackage{interval}
\usepackage{bm}
\usepackage{mathtools}
\usepackage{footmisc}

\usepackage{siunitx}
\usepackage{bbold}
\usepackage{multirow}
\usepackage{bbm}
\usepackage{amssymb}
\usepackage{bbding}
\usepackage[dvipsnames]{xcolor}
\usepackage{makecell}
\usepackage[inline]{enumitem}
\usepackage{algpseudocode,algorithm,algorithmicx}

\usepackage[pagebackref=true,breaklinks=true,letterpaper=true,colorlinks,citecolor=ForestGreen, bookmarks=false]{hyperref}

\newcommand*{\system}{BEVT\@\xspace}
\newcommand{\ssv}{{\scshape SSv2}\xspace}
\newcommand{\diving}{{\scshape Diving\-48}\xspace}
\newcommand{\kn}{{\scshape K400}\xspace}
\newcommand{\inet}{{\scshape ImageNet}\xspace}

\newcommand{\Drop}[1]{\textcolor{red}{\xspace\small{\bf $\downarrow$#1}}}
\newcommand{\ra}[1]{\renewcommand{\arraystretch}{#1}}

\usepackage[capitalize]{cleveref}
\crefname{section}{Sec.}{Secs.}
\Crefname{section}{Section}{Sections}
\Crefname{table}{Table}{Tables}
\crefname{table}{Tab.}{Tabs.}

\def\confName{CVPR}
\def\confYear{2022}

\begin{document}

\title{BEVT: BERT Pretraining of Video Transformers}

\author{Rui Wang$^{1}$\footnotemark[1] \quad Dongdong Chen$^{2}$ \quad Zuxuan Wu$^{1}$\footnotemark[2] \quad Yinpeng Chen$^{2}$ \quad Xiyang Dai$^{2}$\\ \quad Mengchen Liu$^{2}$ \quad Yu-Gang Jiang$^{1}$\footnotemark[2] \quad Luowei Zhou$^{2}$ \quad Lu Yuan$^{2}$\\
\normalsize$^{1}$Shanghai Key Lab of Intelligent Information Processing, \\ \normalsize School of Computer Science, Fudan Univeristy\\
\normalsize$^{2}$Microsoft Cloud + AI
}
\maketitle

\renewcommand{\thefootnote}{\fnsymbol{footnote}}
\footnotetext[1]{Work done during an internship at Microsoft}
\footnotetext[2]{Corresponding authors}

\begin{abstract}

This paper studies the BERT pretraining of video transformers. It is a straightforward but worth-studying extension given the recent success from BERT pretraining of image transformers. We introduce BEVT which decouples video representation learning into spatial representation learning and temporal dynamics learning. In particular, BEVT first performs masked image modeling on image data, and then conducts masked image modeling jointly with masked video modeling on video data. This design is motivated by two observations: 
1) transformers learned on image datasets provide decent spatial priors that can ease the learning of video transformers, which are often times computationally-intensive if trained from scratch; 2) discriminative clues, i.e., spatial and temporal information, needed to make correct predictions vary among different videos  due to large intra-class and inter-class variations. We conduct extensive experiments on three challenging video benchmarks where BEVT achieves very promising results. On Kinetics 400, for which recognition mostly relies on discriminative spatial representations, BEVT achieves comparable results to strong supervised baselines. On Something-Something-V2 and Diving 48, which contain videos relying on temporal dynamics, BEVT outperforms by clear margins all alternative baselines and achieves state-of-the-art performance with a 71.4\% and 87.2\% Top-1 accuracy respectively. Code will be made available at \url{https://github.com/xyzforever/BEVT}.
   
\end{abstract}

\section{Introduction}
\label{sec:intro}

\begin{figure}[t]
\begin{center}
   \includegraphics[width=\linewidth]{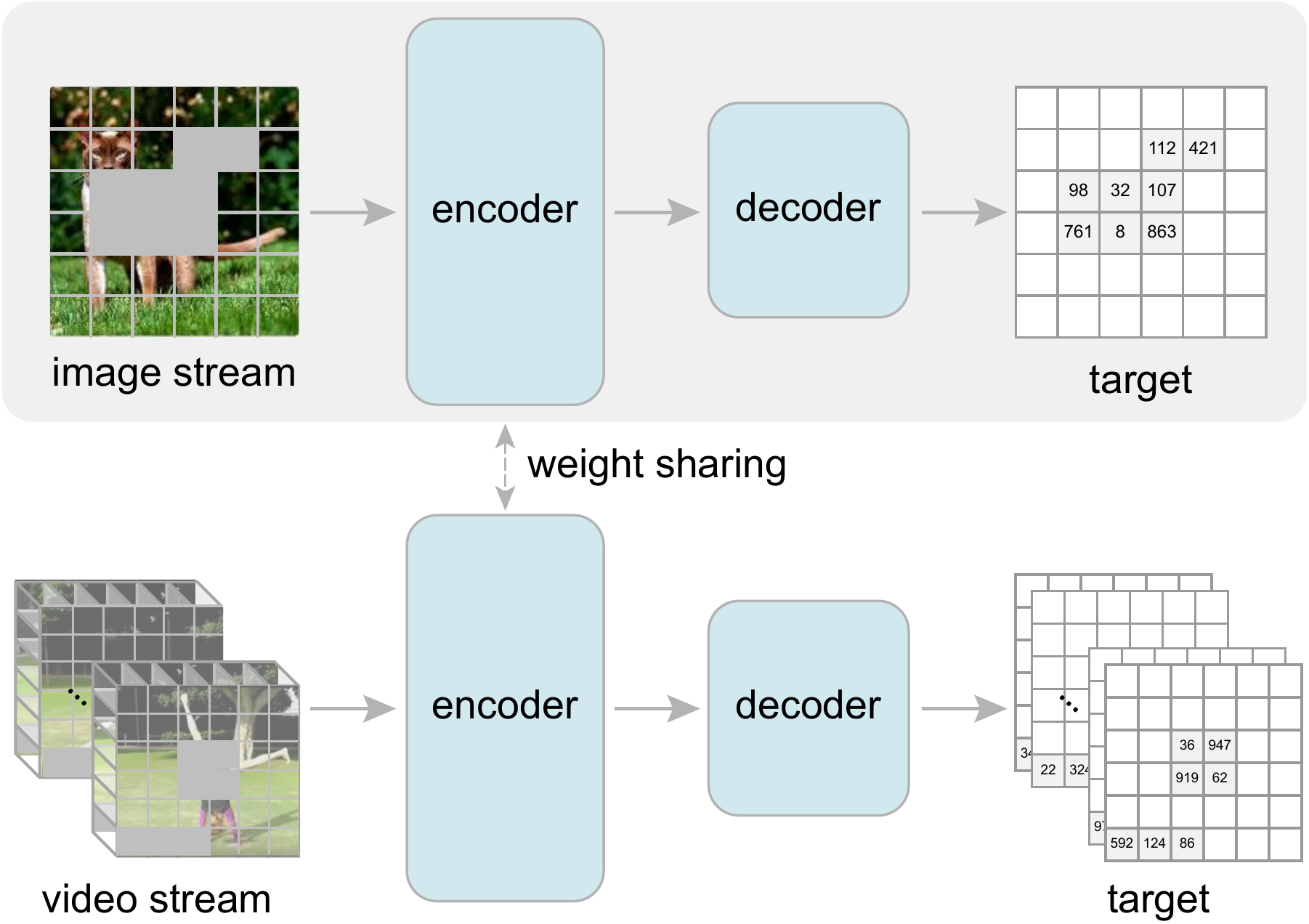}
\end{center}
   \vspace{-0.2in}
   \caption{A conceptual overview of \system. It has a decoupled design: \textbf{first} conducts masked image modeling on image data, and \textbf{then} conducts jointly masked image modeling and masked video modeling on image\&video data by weight sharing.}
\label{fig:decoder}
\end{figure}

Transformers\cite{vaswani2017attention,wolf2020transformers}  have become the dominant network structures in the natural language processing (NLP) field and made tremendous success in different NLP tasks. Recently, the pioneering work ViT\cite{dosovitskiy2020image} proposes to tokenize one image into a series of patch-based tokens and apply the transformer architecture for image recognition.  Many approaches \cite{chen2021mobile,dong2021cswin,liu2021swin,wu2021cvt} further demonstrate the power of transformers as generic vision backbones and achieve state-of-the-art performance on various vision tasks. Beyond image tasks, there are also a few studies showing the promise of transformers for video understanding~\cite{liu2021video,arnab2021vivit}.

The key to the success of Transformers in NLP is BERT pretraining~\cite{devlin2018bert,liu2019roberta,bao2020unilmv2}, one of the most successful pretraining tasks, which predicts masked tokens in corrupted texts. This motivates a few recent studies to explore the BERT-style pretraining for image representation learning by recovering raw pixels~\cite{he2021masked} or latent codes~\cite{bao2021beit,dong2021peco} of masked image patches. However, how to leverage such a training strategy for video understanding has never been explored before.

In this paper, we study BERT pretraining of video transformers. Unlike static images, videos depict how objects move and interact over time. Such dynamic nature brings additional difficulty for representation learning.  It is often found that learning representations from scratch on videos is computationally expensive and requires extremely large-scale datasets with millions of samples ~\cite{largscale2021}, if not hundreds of millions of samples~\cite{akbari2021vatt}. Instead of training from scratch, a few methods demonstrate that self-supervised models pretrained on image datasets benefit video recognition under both supervised~\cite{liu2021video,arnab2021vivit} and unsupervised settings~\cite{gberta_2021_ICML}.  These approaches simply leverage pretrained models as better initializations to learn spatial-temporal features in videos. While widely used and sometimes effective, the spatial context relationships learned from the image prertaining phase are likely to be drastically modified during video feature learning.  

We argue that spatial priors encoded in pretrained self-supervised models should be explicitly preserved when performing video representation learning. The intuition behind is that there are large inter-class variations among different videos and their dependencies on what discriminative information to use (\ie, spatial and temporal clues) to make correct predictions differ. For instance, for actions like ``applying lipstick'', spatial knowledge is generally sufficient, as evidenced by the fact simply using 2D features offers decent results on datasets like Kinetics~\cite{quovadis}. On the other hand, temporal dynamics are crucial for differentiating actions between two fine-grained diving sequences~\cite{resound}. This highlights the importance of considering the differences among video samples during feature learning. 

In light of this, we introduce \system, which decouples video representation learning into spatial representation learning and temporal dynamics learning.  More specifically, \system builds upon the Video Swin Transformer~\cite{liu2021video} due to their computationally efficient architectures~\footnote{Note that we only use the architecture and do not load the pretrained weights.}, and is trained with a BERT-style objective to fully unleash the power of transformers for representation learning. \system contains an image stream for spatial modeling and a video stream for temporal modeling, interacting with each other for video modeling. In particular, the image stream, operating on RGB images, learns spatial priors first on ImageNet in an unsupervised fashion by predicting masked image patches in the form of latent codes derived from a pretrained VQ-VAE as in~\cite{bao2021beit}. It is then used to initialize the attention weight matrices of the video stream, whose inputs are sampled video clips, so as to save computation for video transformers. The video stream, on the other hand, learns temporal dynamics in videos through predicting masked 3D tubes represented by latent codes. The two streams, taking image and video pairs as inputs, are then jointly trained on video data through a weight sharing strategy. Such a design not only maintains spatial knowledge learned from image datasets to ensure decent results for static video samples but also learns temporal information to guarantee correct predictions for samples that contain dynamic movements. Finally, \system is finetuned on targeted datasets for downstream evaluation.

We conduct extensive experiments on three challenging video datasets, \ie, Kinetics-400 (\kn)~\cite{quovadis}, Something-Something-v2 (\ssv)~\cite{ssv2}, and Diving-48 (\diving)~\cite{resound}. On \kn, \system offers $81.1\%$ Top-1 accuracy, which is better than the strong supervised baseline $80.6\%$~\cite{liu2021video}. On \ssv and \diving, \system achieves $71.4\%$ and $87.2\%$ Top-1 accuracy outperforming state-of-the-art methods~\cite{gberta_2021_ICML,liu2021video,arnab2021vivit,slowfast}
by clear margins. To further analyze the performance difference among these three datasets, we further provide the temporal dependency analysis and demonstrate that videos in \kn mainly rely on spatial clues for correct predictions while videos from \ssv and \diving require more temporal information. 

Our main contributions are summarized as follows:
\begin{enumerate*}[label=(\arabic*)]
    \item We explore the BERT-style training  objective to fully unleash the power of transformers to learn discriminative video representations;
    \item We introduce a novel two-stream network that decouples spatial representation learning and temporal dynamics learning; 
    \item We demonstrate different video samples have different preferences towards spatial and temporal clues;
    \item We conduct extensive experiments on three challenging video benchmarks and achieve comparable or better results with state-of-the-art methods. 
\end{enumerate*}

\section{Related Work}
\label{sec:relawork}
\noindent \textbf{Video understanding with CNNs.} There is a plethora of work on video understanding with CNNs, most of which focus on learning spatial-temporal features ~\cite{quovadis,c3d,r21d,slowfast,x3d,tsn,tsm}. These approaches can be divided into two categories: (1) temporal aggregation and (2) 3D CNNs. In particular, temporal aggregation methods typically extract image features/scores frame-by-frame and then combine frame-level information to achieve video-level predictions through recurrent networks~\cite{recurrentdonahue,recurrentjoe} or average pooling~\cite{twostream,tsn}.  On the other hand, 3D CNNs extend 2D convolutions into the time domain by using 3D convolutions on stacked RGB frames for the joint-modeling of spatial-temporal relationships~\cite{quovadis,c3d,r21d,slowfast,x3d}. 3D CNNs are generally computationally expensive, and this motivates a line of research on efficient video recognition~\cite{eco,multifiber,r21d,x3d,trn,tsm,channelseparated}.  Instead of using CNNs, we explore transformers for video understanding due to their strong results on image recognition tasks.

\vspace{0.05in}
\noindent \textbf{Vision transformers.} Motivated by the impressive performance of transformers that are able to capture long-range dependencies in a wide range of natural language processing tasks, there is a growing interest in using transformers for computer vision tasks~\cite{parmar2018image,zhou2018end,dosovitskiy2020image,liu2021swin,fan2021multiscale,deit}. More specifically,  ViT~\cite{dosovitskiy2020image} is the first work that generalizes transformers to the image domain by splitting images into patches that are further embedded with a linear layer as inputs to a Transformer. While demonstrating great potential in image recognition tasks, ViT relies on pretraining on substantially large-scale datasets like ImageNet-21K and the training process is computationally expensive. To mitigate these issues, extensive studies have been introduced. For example, DeiT~\cite{deit} uses a distillation loss to speed up the training process. PiT~\cite{DBLP:conf/cvpr/HanYHY21} incorporates the pooling-based design into transformers. CSwin Transformer \cite{dong2021cswin} proposes multi-head grouping and performs attention within cross-shaped windows. And the recent work Mobile-Former \cite{chen2021mobile} further extends transformer at low FLOP regime. There are also a few very recent studies that extend image transformers for video understanding tasks~\cite{fan2021multiscale,arnab2021vivit,liu2021video}. Fan \etal use a multi-scale design to generate spatial-temporal tokens in different sizes for action recognition~\cite{fan2021multiscale}. Liu \etal extend Swin Transformers into the video domain~\cite{liu2021video}. In this paper, we focus on studying BERT pretraining of video transformer in a self-supervised manner, which is orthogonal to such transformer design efforts.

 \begin{figure*}[t]
\begin{center}
   \includegraphics[width=\linewidth]{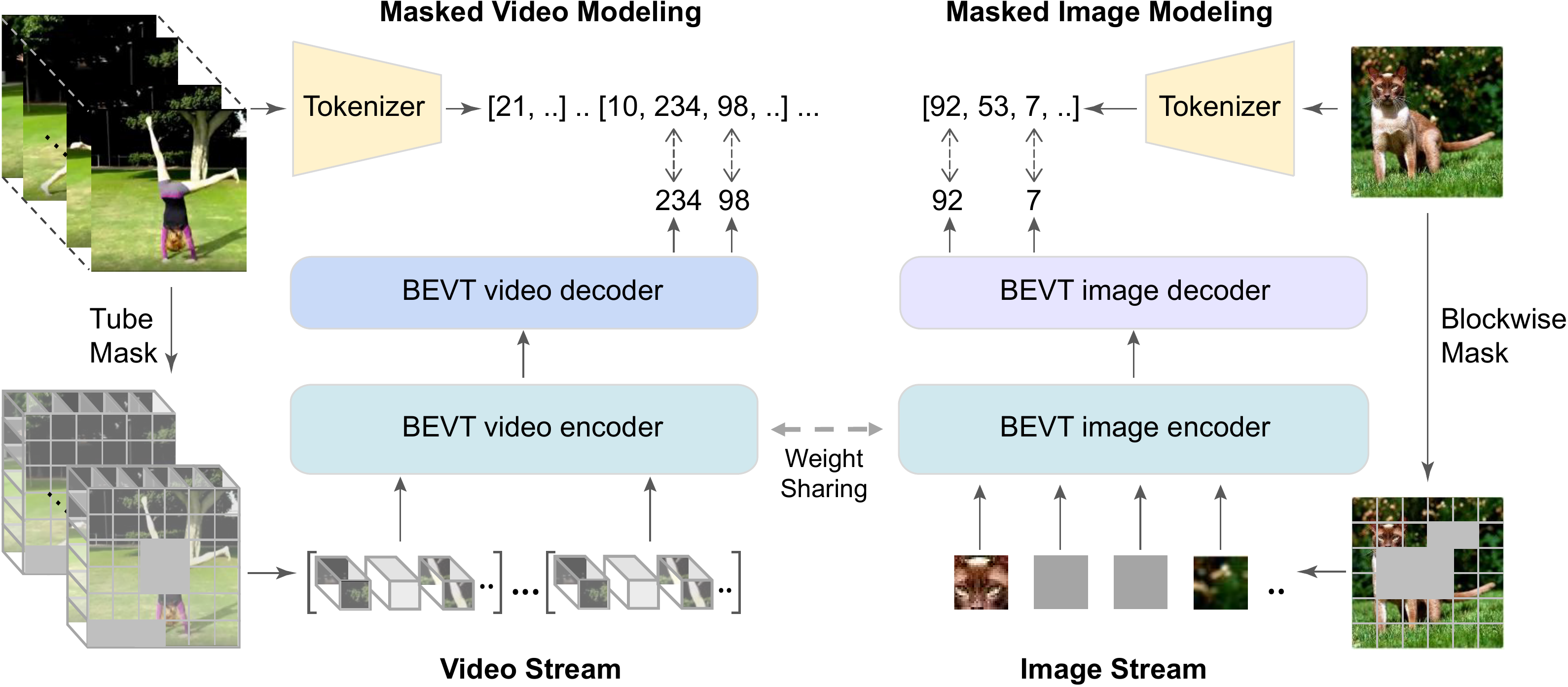}
\end{center}
   \vspace{-0.2in}
   \caption{An overview of our framework. \system contains an image stream and a video stream that learns video representations jointly using a BERT-style objective. In particular, the image and video stream, operating on single images and video cubes, respectively, predict masked image patches and 3D cubes derived from a tokenizer. }
\label{fig:framework}
\end{figure*}

\vspace{0.05in}
\noindent\textbf{Self-supervised representation learning.} At the core of many computer vision tasks is how to learn discriminative features for targeted datasets. Since collecting labeled datasets is labor-intensive and costly, there is an ever-increasing trend in learning representations in a self-supervised manner~\cite{caron2018deepcluster,donahue2019bigbigan,caron2020swav,2020byol,li2021pcl,zbontar2021barlowtwins}.  The main idea is to design surrogate tasks including
 inpainting~\cite{pathak2016inpainting}, colorization~\cite{zhang2016colorful,zhang2017split}, jigsaw predictions~\cite{noroozi2016jigsaw}, rotation predictions~\cite{gidaris2018rotnet}, \etc,  as a form of supervisory signals in lieu of manual labels. More recently, contrastive learning has been a popular paradigm for feature learning by forcing images to be closer to their augmented copies than other samples ~\cite{wu2018npid,ye2019e2e,he2019moco,misra2020pirl,chen2020simclr}. In contrast to these approaches using CNNs as backbones, there are a few very recent studies leveraging contrastive learning~\cite{caron2021emerging,chen2021empirical} for transformers. 
 
\vspace{0.05in}
\noindent\textbf{BERT pretraining.} In contrast to contrastive learning widely used in vision, BERT pretraining \cite{devlin2018bert} is extremely popular and extensively studied \cite{liu2019roberta,bao2020unilmv2} in NLP. As an effort that unifies vision and NLP under the same BERT pretraining framework, the recent work BEiT \cite{bao2021beit} and ICT \cite{wan2021high} utilizes the masked image modeling task to do BERT pretraining of image transformers and achieves great success for different tasks. And one concurrent work PeCo \cite{dong2021peco} further proposes a perceptual codebook to improve the performance. Another concurrent work \cite{he2021masked} extends it from recovering patch tokens to raw pixels. In this paper, we study BERT pretraining for video transformers as an orthogonal unifying effort. Different from BERT pretraining of image transformers and the concurrent effort \cite{tan2021vimpac}, we decouple video pretraining into spatial representation learning and temporal dynamics learning so as to accommodate the varying need of distinct salient clues for different videos.

\section{Method}
The goal of \system is to learn video representations effectively for both relatively static videos and dynamic videos in a self-supervised manner. Here, ``relatively static videos" mean the videos only requiring discriminative spatial representation for recognition, while ``dynamic videos" mean that videos that also require temporal dynamics for recognition. Besides the effectiveness, another key problem to consider in video pretraining is efficiency. Compared to image pretraining, video pretraining is more computationally expensive, thus making pretraining on large-scale video data from scratch inefficient or even inapplicable without massive computational resources.

To this end, \system decouples the video pretraining into spatial representation learning and temporal dynamics learning. And the spatial representation learning is only conducted on image data, while the temporal dynamics learning is conducted on video data. To implement this idea, our \system contains two streams, operating on images and videos, respectively. In the following, we introduce different components of our framework. Figure~\ref{fig:framework} gives an overview of our framework.

\vspace{0.05in}
\noindent\textbf{Image and video patches.} For the video stream, given a video clip $X_{vid} \in \mathbb{R}^{T \times H \times W \times 3} $ with $T$ frames,  we follow VideoSwin \cite{liu2021video} and convert it into $\frac{T}{2} \times \frac{H}{4} \times \frac{W}{4}$ 3D patches, each with a size of $2 \times 4 \times 4 \times 3$; each 3D patch contains a 96-dimensional features.  For the image stream, given an input image $X_{img} \in \mathbb{R}^{H \times W \times 3} $, we consider each patch with a size of $4 \times 4 \times 3$ as a token and set the feature dimension of each token as $48$.  We then project each token to a token embedding vector of dimension $C$ by a linear embedding layer. Then the sequence of token embedding is input into the following transformer architectures.

\vspace{0.05in}
\noindent\textbf{Masked image and video tokens.} Motivated by the great success of BERT in NLP tasks, \system is optimized to simultaneously perform masked image modeling (MIM) and masked video modeling (MVM) by predicting ``corrupted'' image and video tokens, respectively. The MIM is designed to capture spatial priors while the MVM is used to capture temporal dynamics in videos. In particular, for the image stream, since input images are divided into non-overlapping patches, we randomly mask several patches and the image stream is trained to recover them as in~\cite{bao2021beit}. More specifically, the embedded feature of each masked patch is replaced by a learnable mask token embedding. For the video stream, we randomly mask 3D tokens and train the video stream to predict those masked tokens. The set of masked image and video tokens and the remaining patch features are sent to encoders, as will be introduced below.

\vspace{0.05in}
\noindent\textbf{Mask strategy.} For masked image modeling, following ~\cite{bao2021beit}, we use blockwise masking instead of randomly selecting each masked patch. When generating masked positions for an image, we mask a block of patches each time and set the minimum number of patches for each block. The position, the aspect ratio and the size of each block are randomly selected under a preset range. We repeat masking blocks until the ratio of masked patches exceeds the preset lower bound. For masked video modeling, we employ a tube masking strategy that is a straightforward extension of blockwise masking. Given an input video clip of length $T$, we first randomly choose the number of masked frames (tube length) $l$ and the start frame $t$. Then we employ blockwise masking to generate a 2D mask, and apply this 2D mask to each frame from $t$ to $t + l$. In other words, for each masked frame, the set of masked positions is the same and the shape of the whole 3D mask is a tube. The range of the masked tube length is $[0.5T, T]$ and the masking ratio of each masked frame is $0.5$.

\vspace{0.05in}
\noindent\textbf{\system encoders.}  \system contains two encoders, one for the image stream and one for the video stream. Both encoders are instantiated with the Video Swin Transformer~\cite{liu2021video} due to its strong performance with a moderate computational cost. Note that in contrast to~\cite{liu2021video} that performs fully-supervised training, we use the Video Swin Transformer as our backbone for self-supervised learning. In particular, Video Swin Transformer~\cite{liu2021video} follows the design of Swin Transformer~\cite{liu2021swin} and is a hierarchical architecture consisting of four stages. Between every two stages, spatial downsampling is performed by patch merging layers, which concatenates the features of each group of  $2 \times 2$ spatially neighboring patches. After downsampling, a linear layer maps the features of each concatenated token to half of their dimension. A series of Swin attention blocks comes after to apply feature transformation.  

Given a sequence of tokens as inputs, the video encoder outputs a feature map with the size of $\frac{T}{2} \times \frac{H}{32} \times \frac{W}{32} \times 8C$. Since Video Swin Transformer only performs temporal downsampling in the beginning linear embedding layer, it degrades to a 2D architecture when the temporal dimension of the input is 1. As a result, for the image encoder, the output feature map has a size of $\frac{H}{32} \times \frac{W}{32} \times 8C$.

\vspace{0.05in}
\noindent\textbf{Tokenizer.} Following~\cite{bao2021beit}, we use the visual tokens generated by a pretrained image VQ-VAE~\cite{dalle} as the groundtruth tokens and our pretraining task is to predict the tokens for masked patches. The pretrained VQ-VAE tokenizer maps image patch into discrete tokens $z$ by searching the closest latent codes in its pre-learnt visual codebook. Given an input image $X_{img} \in \mathbb{R}^{H \times W \times 3} $, it will be tokenized into the visual token map $Z_{img} \in V^{\frac{H}{16} \times \frac{W}{16}}$. Similarly for an input video  $X_{vid} \in \mathbb{R}^{T \times H \times W \times 3} $,  it is tokenized into visual token map $Z_{vid} \in V^{T \times \frac{H}{16} \times \frac{W}{16}}$. Note that, considering the pretrained VQ-VAE only downsamples $8\times 8$ patch into one token, we downsample the input images/frames by $1/2$ before feeding into the tokenizer so that the output token map has the spatial resolution of $\frac{H}{16}\times\frac{W}{16}$.

\vspace{0.05in}
\noindent\textbf{\system decoders.} To learn meaningful representations by predicting the tokens for the masked image and video patches in inputs, \system has an image decoder and a video decoder as the auxiliary prediction heads, which will be discarded in finetuning stage. 
Existing modern vision transformers including Swin Transformer follow the hierarchical design and downsample the input into decreased spatial/temporal resolutions. Taking the VideoSwin of video stream shown in Figure \ref{fig:decoder} as an example, it consists of four stages, and the feature maps $F_4$ in the last stage have the dimension of $\frac{T}{2}\times\frac{H}{32}\times\frac{W}{32}$. In order to match the dimension of feature maps to the number of groundtruth visual tokens, we design a lightweight decoder for the video stream in BEVT. As shown in Figure \ref{fig:decoder}, it first spatially upsamples the stage-4 feature $F_4$ by using a transposed convolutional layer, and then concatenate the upsampled stage-4 feature $\hat{F_4}$ with stage-3 feature $F_3$ together and fuse them with a simple linear layer. Finally, the fused feature $F$ will be temporally upsampled with another transposed convolution layer.  

\begin{equation}
    \begin{split}
        &\hat{F}_4 = \texttt{Spatial-Upsample}(F_4) \\
        &F = \texttt{Linear}(\texttt{Concat}(\hat{F}_4, F_3)) \\
        &\hat{F} = \texttt{Temporal-Upsample}(F) \\
    \end{split}
\end{equation}

To predict the token for each position $(t, i, j)$, a simple softmax based classifier is applied upon $\hat{F}$:
\begin{equation}
    \mathbf{p}_{t, i, j} = \texttt{softmax}(W f_{t, i, j} + b)
\end{equation}
where $f_{t, i, j}$ is the feature vector of the output feature map $\hat{F}$ at position $(t, i, j)$, $\mathbf{p}_{t,i,j}$ denotes the corresponding probability vector. $W$ and $b$ are the weight and the bias of a linear layer. For the decoder in the image stream, it follows a similar design, and the only difference is without the temporally upsampling part.

\begin{figure}[t]
\begin{center}
   \includegraphics[width=\linewidth]{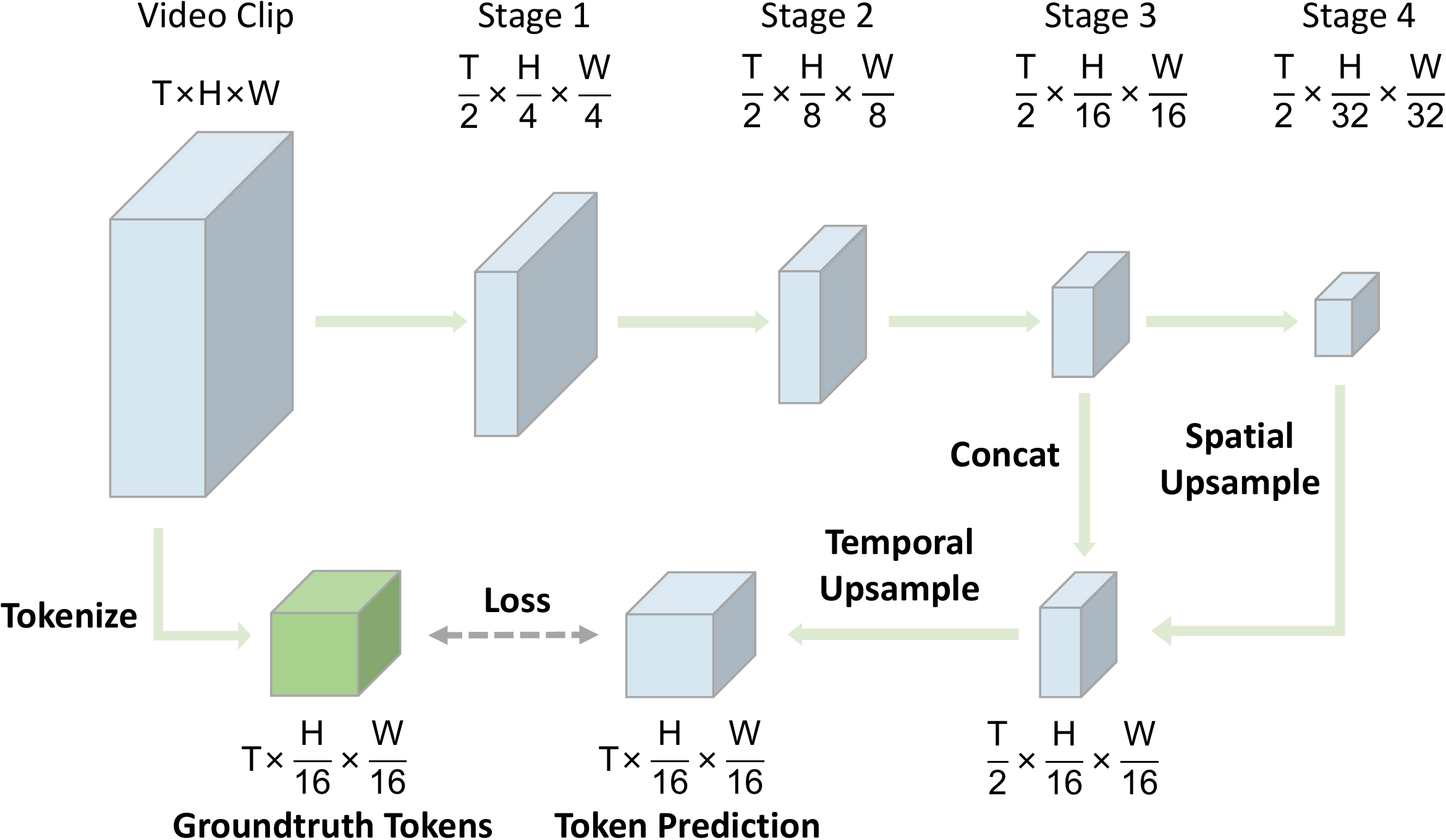}
\end{center}
   \vspace{-0.2in}
   \caption{\system encoder and decoder for masked video modeling. }
\label{fig:decoder}
\end{figure}

\vspace{0.05in}
\noindent\textbf{Training objectives.}  Denote the positions of masked patches in the input images and videos as $M_I$ and $M_V$, the objective of masked image modeling is to maximize the log-likelihood of the groundtruth token $z_{i,j}$ for each mask position $(i,j)$.

\begin{equation}
  L_{MIM} = -\frac{1}{\lvert M_I \rvert} \sum_{(i, j) \in M_I} \log \mathbf{p}_{i,j}^{z_{i,j}}
  \label{eq:mim_loss}
\end{equation}

where the superscript of $\mathbf{p}$ denotes indexing the probability value of one specific position.  Similarly, the target of masked video modeling can be denoted as:
\begin{equation}
  L_{MVM} = - \frac{1}{\lvert M_V \rvert} \sum_{(t, i, j) \in M_V} \log {\mathbf{p}^{z_{t, i, j}}_{t,i,j}}
  \label{eq:mvm_loss}
\end{equation}

The objective of two-stream joint training is a simple combination of two objectives:
\begin{equation}
  L =  L_{MIM} + \lambda L_{MVM}
  \label{eq:joint_loss}
\end{equation}
where $\lambda$ is the hyper-parameter that balances the weights of the image stream and the video stream.

\vspace{0.05in}
\noindent\textbf{Training strategies.}  Following our decoupled design, we \textit{first train the image stream} on ImageNet with the masked image modeling task to learn discriminative spatial representation. 
The resulting model is then used to initialize the video stream, and \textit{both streams are jointly trained} by optimizing Equation~\ref{eq:joint_loss} such that the objective $L_{MIM}$ preserves spatial information while $L_{MVM}$ learns to capture temporal dynamics in videos. Such a strategy not only makes \system much more efficient than pretraining video transformers on large-scale video data from scratch, but also satisfies the need of learning different discriminative clues for different types of video samples. 

\vspace{0.05in}
\noindent\textbf{Weight sharing between streams.} When jointly training the image and video stream, instead of learning two sets of model weights for the two streams independently, we design a weight sharing strategy so that they can share model weights for the encoder except some image/video specific parts. This is motivated by the good property of transformer networks, \ie, most operators (including multi-head attention and FFN) are oriented to tokens but not specific input types. Taking the Video Swin transformer as an example, we use the following strategies for weight sharing: (1) We use independent 2D patch partitioning layers instead of 3D patch partitioning, and add a linear embedding layer in the first stage for projecting image tokens to the same dimension as the original 3D video patch embedding; (2) We adapt 3D shifted local window to the 2D scenario. This is fulfilled by reusing the submatrix of the original 3D relative positional embedding where the relative temporal distance is $0$ as the 2D relative positional embedding. With such a design, the image stream and the video stream can help each other by optimizing one ``mostly-unified" encoder.

\vspace{0.05in}
\noindent\textbf{Finetuning and inference.} Once pretrained, \system provides decent video representations that can be transferred for downstream tasks. On targeted datasets, we simply use the 3D patch embedding layers and the video encoder, to which a few task-specific layers (e.g., classification head for video recognition) are appended, for finetuning. The resulting model can then be readily used for inference.

\section{Experiments}
\subsection{Experimental Setup}
\noindent\textbf{Datasets and evaluation metrics.} We evaluate our method on three representative video recognition datasets: Kinetics-400 (\kn)~\cite{quovadis}, Something-Something-v2 (\ssv)~\cite{ssv2}, and Diving-48 (\diving)~\cite{resound}. \kn contains videos clips from YouTube with an average duration of 10 seconds and the videos are manually labeled  into 400 categories. Following~\cite{slowfast}, we use ${\sim}240K$ videos for training and ${\sim}20K$ videos for testing. \ssv is also a large-scale video dataset that contains ${\sim}160K$ videos for training and ${\sim}20K$ videos for testing. The videos in \ssv are labeled into 174 classes the average duration is 4 seconds. \diving contains ${\sim}17K$ fine-grained diving sequences, which are further split into a training set with around ${\sim}15K$ clips and a testing set with ${\sim}2K$ clips. Compared to \kn, recognizing videos in \ssv and \diving requires more temporal information, as will be introduced below. Following official instructions, we report Top-1 accuracy on all three datasets. And the default resolution $224\times 224$ is used.

\vspace{0.05in}
\noindent\textbf{Implementation Details.} We use Video Swin-Base for experiments throughout the paper unless mentioned otherwise. For pretraining the image stream \system-I alone, we train the model for 800 epochs on  ImageNet-1K with a batch size of 2048. For pretraining the video stream \system-V alone or the two-stream \system, we train the model for 150 epochs on \kn with a batch size of 256 and the clip length $T$ is 16. When performing two stream pretraining, \inet images are employed to train the image stream with a batch size of 2048, and the loss weight $\lambda$ is simply set to $1$. We use the DALL-E tokenizer~\cite{dalle} unless explicitly stated. For downstream tasks, we finetune the pretrained models for 60 epochs with a batch size of 64 and the clip length is 32. We use the AdamW~\cite{loshchilov2018decoupled} optimizer with a linear warm-up and a cosine learning rate schedule for both pretraining and finetuning. The pretraining on the image stream takes about 8 days on 16 V100 GPUs. The two-stream joint pretraining of 150 epochs takes about 4 days on 32 V100 GPUs.

\subsection{Main Results}
\noindent\textbf{Effectiveness of \system for video transformer pretraining.} To demonstrate the effectiveness of \system, we compare it with the four image transformer pretraining baselines:
\begin{enumerate*}[label=(\arabic*)]
    \item \emph{Image Sup}: pretraining the image Swin Transformer on Imagenet-1K in a supervised way. Similar strategies are commonly used in existing video transformer papers \cite{liu2021video,arnab2021vivit,gberta_2021_ICML}. 
    \item \emph{Image CL}: pretraining the image Swin Transformer on Imagenet-1K with a self-supervised contrastive learning method \cite{xie2021self}.
    \item \emph{BEVT-I}: pretraining the image Swin Transformer only with the image stream, which is similar to BEiT \cite{bao2021beit}.
    \item \emph{BEVT-V}: pretraining the video Swin Transformer only with the video stream.
\end{enumerate*}
 The pretrained weights from \emph{Image Sup}, \emph{Image CL} and \emph{BEVT-I} used as the initialization of the Video Swin Transformer for finetuing. For the video stream, we design two baselines by conducting BERT pretraining on K400 and HowTo100M~\cite{miech2019howto100m} from scratch, \ie, \emph{BEVT-V}, which is our framework without the decoupled design. As emphasized before, because video pretraining is more computationly expensive than image pretraining, pretraining on HowTo100M dataset with many epochs is not applicable. For fair comparisons, we also use 32 V100 GPUs to pretrain HowTo100M about 8 days (about 2 epochs). 

The comparison results are summarized in Table~\ref{tab:pretrain}. We observe that: 
\begin{enumerate*}[label=(\arabic*)]
\item \system outperforms the \emph{Image Sup} baseline by clear margins (4.3\% and 2.7\%) on \ssv and \diving, respectively. This not only suggests that learning representations with \system using a BERT-style training objective is promising without the need for manual labels, but also shows that only image-based pretraining is not enough for these two datasets. On \kn, the performance of \system is on par with the \emph{Image Sup} baseline. 
\item We also see that \system offers comparable or better results compared to \emph{Image CL} on these three datasets. 
\item Compared to \system-I, \system is better on \ssv and \diving by 1.4\% and 5.5\% respectively, highlighting the gains brought by the video stream. Similarly, \system obtains similar results as \system-I on \kn.
\item Compared to \system, \system-V pretraining on \kn or HowTo100M from scratch under the similar computation budget achieves much worse results. We hypothesize it may be because the data diversity of K400 is not as good as ImageNet. And for HowTo100M, pretraining much more epochs may help learn better video representation, but it is too costly. This also further justifies the decoupled design in our \system. 
\end{enumerate*}

\begin{table}[t]
  \centering
  \setlength{\tabcolsep}{0pt} 
  \ra{1.0}
  \begin{tabular*}{\linewidth}{@{\extracolsep{\fill}}llccc@{}}
    \toprule
    Method & Pretrain & \ssv & \diving & \kn \\
    \midrule
    Image Sup & IN-1K & 66.3 & 84.0 & 80.6 \\
    \midrule
    Image CL & IN-1K & 67.1 & 85.5 & \textbf{80.9} \\
    \midrule
    \system-I & IN-1K & 69.2 & 81.2 & 80.5 \\
    \system-V & K400 & 67.1 & 83.7 & 76.2 \\
    \system-V & HowTo100M & 64.2 & 82.3 & 75.1 \\
    \midrule
    \system & IN-1K+K400 & \textbf{70.6} & \textbf{86.7} & 80.6 \\
    \bottomrule
  \end{tabular*}
  \caption{Comparison of different pretraining methods. Video Swin-Base is used here.}
  \label{tab:pretrain}
\end{table}

\begin{table}[t]
  \centering
  \setlength{\tabcolsep}{0pt} 
  \ra{1.0}
  \begin{tabular*}{\linewidth}{@{\extracolsep{\fill}}lccr@{}}
    \toprule
    Dataset & Normal & Single-frame & Random-Shuffling \\
    \midrule
    \kn & 80.6 & 65.3\Drop{15.3} & 77.8\Drop{\,\,\,2.8} \\
    \ssv & 66.3 & \,\,6.3\Drop{60.0} & \,\,19.0\Drop{47.3} \\
    \diving & 84.0 & 13.8\Drop{70.2} & 50.4\Drop{33.6} \\
    \bottomrule
  \end{tabular*}
  \caption{Effects of removing temporal information for different video datasets. Video Swin-Base models trained with labeled video data are used. We show top-1 accuracy for evaluation.}
  \label{tab:tepmporal_analysis}
\end{table}

\vspace{0.05in}
\noindent\textbf{Deeper dataset analysis.} To further understand the performance variations of \system among three datasets, we perform a temporal dependency study to investigate the amount of temporal information required for correct predictions. Specifically, we use the following two testing strategies: (1) Single-frame, where we randomly sample a frame and replace all other frames with this one, leading to a static video; (2) Random-Shuffling, where a random shuffling is performed along the temporal axis. The results are summarized in \ref{tab:tepmporal_analysis}. We observe that both strategies have a relatively small impact on \kn compared to \ssv and \diving where there is a 60\% and 70\% performance drop when using the Single-frame strategy. This suggests most videos in \kn can be recognized by discriminative spatial clues whereas temporal dynamics is particularly important for \ssv and \diving. Comparing across Table~\ref{tab:pretrain} and Table~\ref{tab:tepmporal_analysis}, we make the following conclusions: (1) On datasets like \kn where spatial clues are dominant, finetuning a model with spatial priors, \eg pretrained on \inet, can achieve decent performance. Additional video modeling brings little effect to the overall performance; (2) The use of the video stream in \system is crucial to learn necessary temporal information for datasets like \ssv and \diving. This confirms our hypothesis that different videos rely on different discriminative clues for accurate predictions due to the large intra-class and inter-class variations among videos.

\vspace{0.05in}
\noindent\textbf{Comparisons with State-of-the-art methods.} We compare \system with state-of-the-art methods on \ssv, \diving and \kn. On \ssv and \diving, we see from Table~\ref{tab:ssv2} and Table~\ref{tab:diving48} that our approach achieves the best performance by clear margins when compared to existing SOTA methods, including supervised models. It is worth mentioning that, on \ssv, a common practice to achieve better performance is to perform two rounds of pretraining---a model is  pretrained on both \inet and \kn in a fully supervised fashion before finetuning the model on \ssv. Instead, we pretrain on \inet and \kn without using any manual labels, yet our performance is still better. On \kn, we see from Table~\ref{tab:k400} that \system achieves competitive results with state-of-the-art methods using similar or less computation measured by GFLOPs.

\begin{table}
  \centering
  \ra{1.0}
  \setlength{\tabcolsep}{0pt} 
  \begin{tabular*}{\linewidth}{@{\extracolsep{\fill}}lccc@{}}
    \toprule
    Method & Pretrain & Top-1 & \makecell[c]{GFLOPs \\ $\times$ crops} \\
    \midrule
    TimeSformer-HR~\cite{gberta_2021_ICML} & IN-21K & 62.5 & 1703 $\times$ 3 \\
    SlowFast R101~\cite{slowfast} & K400 & 63.1 & 106 $\times$ 3 \\
    TSM-RGB~\cite{tsm} & K400 & 63.3 & 62 $\times$ 6 \\
    MSNet~\cite{kwon2020motionsqueeze} & IN-21K & 64.7 & 67 $\times$ 1 \\
    blVNet~\cite{fan2019more} & SSv2 & 65.2 & 129 $\times$ 1 \\
    ViViT-L~\cite{arnab2021vivit} & - & 65.4 & 903 $\times$ N/A \\
    MViT-B~\cite{fan2021multiscale} & K400 & 67.7 & 455 $\times$ 3 \\
    Mformer-L~\cite{patrick2021keeping} & IN-21K+K400 & 68.1 & 1185 $\times$ 3 \\
    Swin-B~\cite{liu2021video} & IN-1K & 66.3 & 321 $\times$ 3 \\
    Swin-B~\cite{liu2021video} & IN-21K+K400 & 69.6 & 321 $\times$ 3 \\
    \midrule
    BEVT & IN-1K+K400 & \textbf{70.6} & 321 $\times$ 3 \\
    BEVT $\dagger$ & IN-1K+K400 & \textbf{71.4} & 321 $\times$ 3 \\
    \bottomrule
  \end{tabular*}
  \caption{Comparison to state-of-the-art on \ssv. $\dagger$ denotes that we use the IN-1K pretrained PeCo tokenizer~\cite{dong2021peco} instead of the DALL-E tokenizer~\cite{dalle} during pretraining.}
  \label{tab:ssv2}
  \vspace{-1em}
\end{table}

\begin{table}[]
  \centering
  \ra{1.0}
    \setlength{\tabcolsep}{0pt} 
  \begin{tabular*}{\linewidth}{@{\extracolsep{\fill}}lccc@{}}
    \toprule
    Method & Pretrain & Top-1 & Params \\
    \midrule
    SlowFast R101~\cite{slowfast} & K400 & 77.6 & 53.3M \\
    TimeSformer-L~\cite{gberta_2021_ICML} & IN-21K & 81.0 & 121.4M \\
    TQN~\cite{zhang2021temporal} & K400 & 81.8 & N/A \\
    Swin-B~\cite{liu2021video} & IN-1K & 84.0 & 88.1M \\
    \midrule
    BEVT & IN-1K+K400 & \textbf{86.7} & 88.1M \\
    BEVT $\dagger$ & IN-1K+k400 & \textbf{87.2} & 88.1M \\
    \bottomrule
  \end{tabular*}
  \caption{Comparison to state-of-the-art on \diving. $\dagger$ denotes that we use the IN-1K pretrained PeCo tokenizer~\cite{dong2021peco} instead of the DALL-E tokenizer~\cite{dalle} during pretraining.}
  \label{tab:diving48}
  \vspace{-1em}
\end{table}

\begin{table}
  \centering
\setlength{\tabcolsep}{0pt} 
  \ra{1.0}
  \begin{tabular*}{\linewidth}{@{\extracolsep{\fill}}lccc@{}}
    \toprule
    Method & Pretrain & Top-1 & \makecell[c]{GFLOPs \\ $\times$ crops} \\
    \midrule
    R(2+1)D~\cite{r21d} & - & 72.0 & 75 $\times$ 10 \\
    I3D~\cite{quovadis} & IN-1K & 72.1 & 108 $\times$ N/A \\
    NL I3D-101~\cite{nonlocal} & IN-1K & 77.7 & 359 $\times$ 30 \\
    ip-CSN-152~\cite{tran2019video} & - & 77.8 & 109 $\times$ 30 \\
    SlowFast R101~\cite{slowfast} & - & 79.8 & 234 $\times$ 30 \\
    X3D-XXL~\cite{x3d} & - & 80.4 & 144 $\times$ 30 \\
    \midrule
    MViT-B, 32$\times$3~\cite{fan2021multiscale} & - & 80.2 & 170 $\times$ 5 \\
    MViT-B, 64$\times$3~\cite{fan2021multiscale} & - & 81.2 & 455 $\times$ 9 \\
    Mformer~\cite{patrick2021keeping} & IN-21K & 79.7 & 369.5 $\times$ 30 \\
    ViT-B-VTN~\cite{neimark2021video} & IN-21K & 78.6 & 4218 $\times$ 1 \\
    TimeSformer-L~\cite{gberta_2021_ICML} & IN-21K & 80.7 & 2380 $\times$ 3 \\
    ViViT-L/16$\times$2~\cite{arnab2021vivit} & IN-21K & 80.6 & 1446 $\times$ 12 \\
    Swin-B~\cite{liu2021video} & IN-1K & 80.6 & 282 $\times$ 12 \\
    \midrule
    BEVT & IN-1K+K400 & 80.6 & 282 $\times$ 12 \\
    BEVT $\dagger$ & IN-1K+K400 & 81.1 & 282 $\times$ 12 \\
    \bottomrule
  \end{tabular*}
  \caption{Comparison to state-of-the-art on \kn. $\dagger$ denotes that we use the IN-1K pretrained PeCo tokenizer~\cite{dong2021peco} instead of the DALL-E tokenizer~\cite{dalle} during pretraining.}
  \label{tab:k400}
  \vspace{-1em}
\end{table}

\subsection{Ablation Study}
Below, we provide a set of ablation studies to justify the contribution of different components in our framework.

\vspace{0.05in}
\noindent\textbf{Importance of image stream pretraining.} In our \system, we first conduct the image stream pretraining  only on the large-scale image data to efficiently learn spatial representations, then use it as the initialization for joint pretraining. To show its importance, we provide some ablation results in Table~\ref{tab:init_ablation}, where the column ``Init" means whether to use the image stream pretrained weights as initialization or not. We have some interesting findings: (1) Using the image stream pretrained weights as initialization can benefit both pure video stream pretraining (\ie,``BEVT-V") and the following joint pretraining of image stream and video stream (\ie,``BEVT"). (2) Even with the initializations, jointly training the image stream with the video stream is still necessary and can bring desirable performance gain. 

\vspace{0.05in}
\noindent\textbf{Image data in joint training.} By default, when jointly learning spatial and temporal representations, the image stream in \system continues to use the \inet images as training images. In this ablation, we also experiment with a variant that uses \kn frames for the image stream. The results are shown in Table~\ref{tab:pretrain_ablation}. We see that images from \inet are slightly better than those from \kn, \ie less than 0.3\% on all three datasets. This suggests that the image stream, designed to preserve spatial knowledge, is not very sensitive to data domains.

\begin{table}[t]
  \centering

\ra{1.0}
  \setlength{\tabcolsep}{0pt} 
  \begin{tabular*}{\linewidth}{@{\extracolsep{\fill}}lcccc@{}}
    \toprule
    Method & Init & \ssv & \diving & \kn \\
    \midrule
    BEVT-I & - & 69.2 & 81.2 & 80.5 \\
    \midrule
    BEVT-V  & $\times$ & 67.1 & 83.7 & 76.2 \\
    BEVT-V  & $\checkmark$ & 70.0 & 85.2 & 79.6 \\
    \midrule
    BEVT  & $\times$ & 67.9 & 85.1 & 78.5 \\
    BEVT  & $\checkmark$ & \textbf{70.6} & \textbf{86.7} & \textbf{80.6} \\
    \bottomrule
  \end{tabular*}
  \caption{Ablation study to show the importance of image stream pretraining. Init means models are initialized from image transformers pretrained with the image stream on ImageNet-1K.}
  \label{tab:init_ablation}
  \vspace{-1em}
\end{table}

\begin{table}
  \centering
  \ra{1.0}
    \setlength{\tabcolsep}{0pt} 
  \begin{tabular*}{\linewidth}{@{\extracolsep{\fill}}lccccc@{}}
    \toprule
    Method & Video & Image & SSv2 & Diving48 & K400 \\
    \midrule
    BEVT-V & K400 & - & 70.0 & 85.2 & 79.6 \\
    \midrule
    BEVT & K400 & K400 & 70.3 & 86.6 & 80.5 \\
    BEVT & K400 & IN-1K & \textbf{70.6} & \textbf{86.7} & \textbf{80.6} \\
    \bottomrule
  \end{tabular*}
  \caption{Ablation study on the image data of joint pretraining. Models are initialized from image transformers pretrained with the image stream on ImageNet-1K.}
  \label{tab:pretrain_ablation}
  \vspace{-1em}
\end{table}

\vspace{0.05in}
\noindent\textbf{Different Pretrained Tokenizer.} We also experiment with the PeCo tokenizer~\cite{dong2021peco} instead of the DALL-E tokenizer~\cite{dalle} in  BEVT. PeCo is only pretrained on ImageNet-1K and uses the same codebook size as in DALL-E. 
As the results shown in Table~\ref{tab:ssv2}-\ref{tab:k400}, PeCo tokenizer outperforms the DALL-E tokenizer on all three datasets and pushes BEVT to higher state-of-the-art performance with 71.4\% and 87.2\% Top-1 accuracy on \ssv and \diving respectively. This demonstrates that better performance can be achieved with a better visual tokenizer.

\vspace{0.05in}
\noindent\textbf{Effect of masking strategies.} We evaluate the \system-V with different masking strategies for the video stream, \ie the temporal length to be masked and the ratio of masking. The experiments are conducted with Video Swin Tiny for time consideration. In addition to tube masking strategies, we compare with:
\begin{enumerate*}[label=(\arabic*)]
    \item \emph{Random-3D}: which samples random patches and masks them following a uniform distribution.
    \item \emph{Frame-Diff}: which uses the same strategy to choose masked frames as tube masking, but applies blockwise masking independently for each frame. 2D masks may be different for different masked frames.
    \item \emph{Random-Frame}: which samples random frames and masks them with the same 2D mask generated by blockwise masking. The temporal positions of masked frames may not be consecutive.
\end{enumerate*}
The results are summarized in Table~\ref{tab:mask_ablation}. We have several observations: 
\begin{enumerate*}[label=(\arabic*)]
\item Masking tubes offers the better results compared to other masking methods like Random-3D and Random-Frame. 
\item Setting too small tube temporal length (e.g., [0.25T, 0.75T]) or too large temporal length (e.g., T) will both incur inferior results on \ssv. We guess it is because the former setting will make the masked video modeling too easy while the later will degrade to masked image model to some extent.(3) Applying different block masks for different frames (``Frame-Diff") is also not good, which possibly shares the similar reason as the small temporal length, \ie, making masked video modeling too easy because information can be easily borrowed from adjacent/short-term frames.
\end{enumerate*}

\begin{table}[t]
  \centering
 \ra{1.0}
    \setlength{\tabcolsep}{0pt} 
  \begin{tabular*}{\linewidth}{@{\extracolsep{\fill}}lcccc@{}}
    \toprule
    Strategy & Length & Ratio & SSv2 & K400 \\
    \midrule
    Tube & 0.5T-T & 40\% & 61.5 & 70.9 \\
    Tube & 0.5T-T & 50\% & 63.3 & 71.6 \\
    Tube & 0.5T-T & 60\% & 63.6 & 71.4 \\
    Tube & 0.5T-T & 70\% & 63.5 & 71.4 \\
    \midrule
    Tube & 0.25T-0.75T & 50\% & 62.8 & 69.5 \\
    Tube & 0.75T-1.0T & 50\% & 63.1 & 71.6 \\
    Tube & T & 50\% & 62.6 & 71.2 \\
    \midrule
    Random-3D & - & 50\% & 59.1 & 67.4 \\
    Frame-Diff & 0.5T-T & 50\% & 62.9 & 70.6 \\
    Random-Frame & 0.5T-T & 50\% & 62.6 & 70.9 \\
    \bottomrule
  \end{tabular*}
  \caption{Ablation study on the mask strategy. Video Swin-Tiny is used for this study.}
  \label{tab:mask_ablation}
  \vspace{-1em}
\end{table}

\section{Conclusion and Discussion}
In the NLP field, transformers have become the de-facto standard architecture and reshaped varieties of NLP tasks in the past several years. This is largely driven by the widely used BERT pretraining strategy that demonstrates scaling abilities for pretraining large models on large-scale data. Recent success of transformers on a variety of computer vision tasks motivates a line of work to explore BERT pretraining in vision.

Unlike a few concurrent studies on images, we take a step forward and study how to explore BERT pretraining for video transformers. This may look very straightforward, but worth being extensively studied. We introduced BEVT that learns both discriminative spatial representation and temporal dynamics. We demonstrate that decoupling video pretraining into spatial and temporal representation learning is not only efficient but also effective. Empowered by the simple design of BEVT, we achieved SOTA performance on three video recognition datasets. We hope our study will inspire more research efforts in this direction.

\vspace{0.05in}
\noindent\textbf{Limitations.} Although our decoupled design has significantly improved the video pretraining efficiency, BEVT still requires pretty high computational resources (\eg, more than one week on 32 V100 GPUs). This would be the biggest obstacle in scaling the pretraining on larger video datasets and larger models. The idea proposed in the concurrent work \cite{he2021masked} may help, \ie, ignoring the masked tokens in the transformer computation during BERT pretraining, which is left as the future work.    
{\small
\bibliographystyle{ieee_fullname}
\bibliography{egbib}
}

\clearpage
\newpage
\appendix

\section{Details on Pretrained Tokenizer}

In this paper, we use the visual tokenizer of a pretrained image VQ-VAE from~\cite{dalle}, which is also called discrete VAE~\cite{dalle}. The tokenizer of the VQ-VAE is trained to transform each $256 \times 256$ image into a $32 \times 32$ image token map according to a visual codebook, while the decoder of the VQ-VAE is trained to reconstruct each input image from its tokens. The vocabulary size of the visual tokens is $8192$. 

\end{document}